\documentclass[10pt,twocolumn,letterpaper]{article}

\usepackage{cvpr}
\usepackage{times}
\usepackage{epsfig}
\usepackage{graphicx}
\usepackage{amsmath}
\usepackage{amssymb}
\usepackage{multirow,slashbox,makecell,array,booktabs}
\usepackage[pagebackref=true,breaklinks=true,letterpaper=true,colorlinks,bookmarks=false]{hyperref}

\cvprfinalcopy % *** Uncomment this line for the final submission

\def\cvprPaperID{1998} % *** Enter the CVPR Paper ID here

\ifcvprfinal\fi
\begin{document}

%%%%%%%%% TITLE
\title{Box-driven Class-wise Region Masking and Filling Rate Guided Loss for \\
Weakly Supervised Semantic Segmentation}

\author{Chunfeng Song$^{1,2}$ \hspace{7mm}  Yan Huang$^{1,2}$ \hspace{7mm} Wanli Ouyang$^{3}$  \hspace{7mm} Liang Wang$^{1,2,4,5}$\\
$^1$Center for Research on Intelligent Perception and Computing (CRIPAC),\\
National Laboratory of Pattern Recognition (NLPR),\\
Institute of Automation, Chinese Academy of Sciences (CASIA)\\
$^2$University of Chinese Academy of Sciences (UCAS)\\
$^3$The University of Sydney, SenseTime Computer Vision Research Group, Australia\\
$^4$Center for Excellence in Brain Science and Intelligence Technology (CEBSIT)\\
$^5$Chinese Academy of Sciences - Artificial Intelligence Research (CAS-AIR)\\
{\tt\small \{chunfeng.song, yhuang, wangliang\}@nlpr.ia.ac.cn \hspace{5mm} wanli.ouyang@sydney.edu.au}
}

\maketitle
\thispagestyle{empty}

%%%%%%%%% ABSTRACT
\begin{abstract}
 Semantic segmentation has achieved huge progress via adopting deep Fully Convolutional Networks (FCN). However, the performance of FCN based models severely rely on the amounts of pixel-level annotations which are expensive and time-consuming. To address this problem, it is a good choice to learn to segment with weak supervision from bounding boxes. How to make full use of the class-level and region-level supervisions from bounding boxes is the critical challenge for the weakly supervised learning task. In this paper, we first introduce a box-driven class-wise masking model (BCM) to remove irrelevant regions of each class. Moreover, based on the pixel-level segment proposal generated from the bounding box supervision, we could calculate the mean filling rates of each class to serve as an important prior cue, then we propose a filling rate guided adaptive loss (FR-Loss) to help the model ignore the wrongly labeled pixels in proposals. Unlike previous methods directly training models with the fixed individual segment proposals, our method can adjust the model learning with global statistical information. Thus it can help reduce the negative impacts from wrongly labeled proposals. We evaluate the proposed method on the challenging PASCAL VOC 2012 benchmark and compare with other methods. Extensive experimental results show that the proposed method is effective and achieves the state-of-the-art results.
\end{abstract}

%%%%%%%%% BODY TEXT
\section{Introduction}

\begin{figure}[t]
	\begin{center}
		\includegraphics[width=1\linewidth]{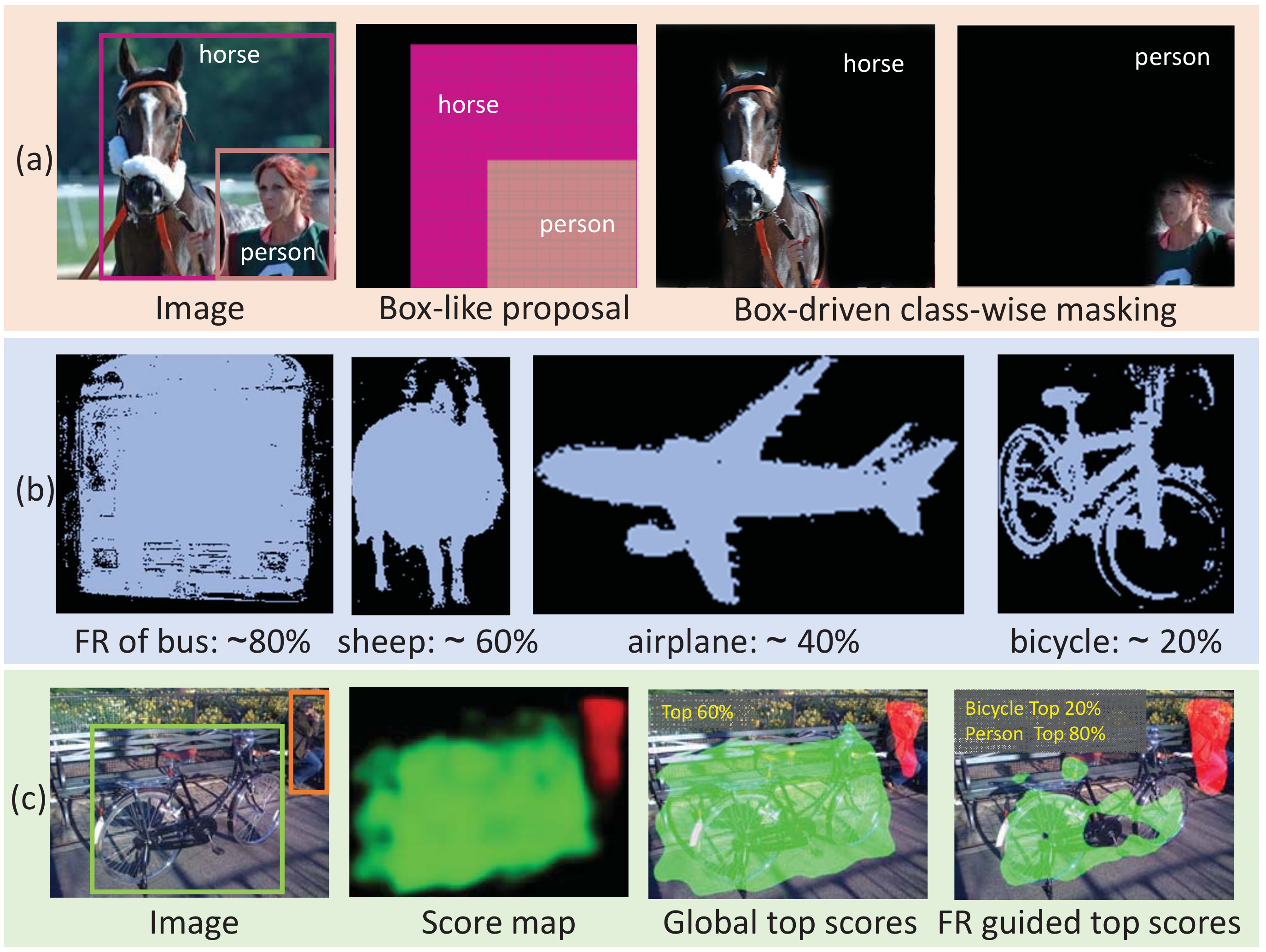}
	\end{center}
	\caption{Weakly supervised segmentation with the box-level annotations. (a) The box-driven class-wise masking (BCM) model can learn specific masks for each class in region-level, and help remove the irrelevant regions of each class softly. (b) Based on the pixel-level segment proposals and the bounding boxes, we could calculate the mean pixel filling rates of each class, e.g., the sheep fills roughly 60$\%$ pixels of the box. (c) Via ranking the values of the score map, we can select the most confident locations for back propagation and ignore the weak ones. As shown in the picture, filling rate guided top scores selection is better than the global one.}
	\label{fig:fig1}
%\vspace{-3.0mm}
\end{figure}

Semantic image segmentation refers to classifying each pixel in an image. Recently, semantic segmentation has achieved a series of progress \cite{R1, R3, R4, R24, R25, R26, R28, R36}, among which \cite{R1} is the first to introduce Fully Convolutional Networks (FCN) structure into segmentation field. Following this work, there are some improvements through redesigning or adjusting the FCN structures \cite{R2, R16, R17, R21, R27, R34}. However, these works are designed for fully supervised mode, which has to be trained with large amounts of fully labeled data. Unlike other classic visual tasks such as classification and object detection, labeling semantic segmentation is rather expensive. For example, the cost of labeling a pixel-level segmentation annotation is about 15 times larger than labeling a bounding box, and 60 times than labeling an image class \cite{R33}. Considering bounding boxes also contain abundant semantic and objective information, a straightforward idea is to learn segmentation weakly with the bounding box supervision.

Recently, several weakly supervised segmentation methods \cite{R5, R6, R7, R35, R37} have been proposed to learn semantic masks with bounding box supervision. These methods mainly focus on generating high-quality pixel-level proposals. For example, in \cite{R6}, the unsupervised dense CRF \cite{R9} was applied to eliminate the background within the bounding box. SDI \cite{R7} tried to produce segment proposals via combining MCG \cite{R10} and GarbCut \cite{R11} methods. BoxSup \cite{R5} updated the candidate masks generated by MCG in an iterative way. Then taking these enhanced segmentation proposals as pixel-level supervision, the deep FCN model can be trained for weakly supervised segmentation. Therefore, it is a core problem how to guide the FCN model to focus on the correct object regions and ignore the wrongly labeled regions from the segment proposals. Most previous approaches train the models with fixed proposals or simple iterative training. In this case, the gap between the ground-truth annotations and generated proposals limits their performance. We address this problem from two aspects.

First, considering that bounding boxes contain strong semantic and objective information, they should help us to remove the irrelevant regions and focus on the foreground regions. A straightforward idea is to learn a global mask to help remove the backgrounds in the images. However, the global mask can not learn multiple accurate shape templates for each class at the same time. To this end, we explore to adopt a box-driven class-wise masking (BCM) model to filter the feature maps of each class with boxes supervision, as shown in Figure \ref{fig:fig1} (a). The learned class-wise masks can provide obvious shape and location hints for each object, which is useful for the following segmentation learning.

Second, filling rate is a useful guidance for obtaining pseudo labels. It is well known that the score map in well trained model has different response values, indicating the confidence of prediction. A natural idea is to select the locations with the most active scores for backward learning, whereas ignore the less confident ones, as shown in Figure \ref{fig:fig1} (c). However, it is difficult to determine the threshold in a weakly supervised task, especially that different classes may need different thresholds. As shown in Figure \ref{fig:fig1} (b), different classes usually has different shapes, e.g., bus has 80$\%$ foreground pixels within its box while bicycle only fills 20$\%$ pixels of the box. This phenomenon inspires us to compute the mean filling rates of each class. Taking the pixel-level segment proposals generated through unsupervised methods as pseudo labels, we could calculate the mean pixel filling rates of each class. We find that the percentage of foreground pixels within the bounding box should be similar for the same class. Whereas the pixel filling ratios of two classes are usually different. Since the segment proposal for single sample is usually not accurate, the mean filling rate for samples of the same class can provide a more stable guidance. Rethinking the discussion above for the mean pixel filling rates, it will be a good choice to guide the top score selection with the filling rate. Based on this motivation, we propose a filling rate guided adaptive loss (FR-loss) to adjust the pseudo labels. Considering the situation that two objects from the same class may have different filling rates due to the shape and pose varieties, we try to refine the filling rates via clustering each class into several sub-classes.

Based on the analysis above, we propose the box-driven class-wise region masking (BCM) model and filling rate guided loss (FR-loss) for weakly supervised semantic segmentation. Firstly, we implement the BCM via segmentation guided learning with a box-like supervision. The proposed BCM can help remove the irrelevant regions of each class softly. It also provides an obvious hint of the foreground region, which could greatly contribute to the segmentation learning. Secondly, we calculate the mean filling rates of each class with the given bounding boxes and the generated pixel-level pseudo proposals. Thus we propose a filling rate loss to help select the most confident locations in the score map for back propagation and ignore the wrongly labeled pixels in proposals. With BCM and FR-loss working together, we could achieve the best performance with weak box supervision. We evaluate the proposed method on the challenging PASCAL VOC2012 dataset \cite{R14} and compare with previous methods. Extensive experimental results demonstrate that the our method is effective and achieves the state-of-the-art results. The performance of proposed method is even comparable with the fully supervised model. We summarize our contributions as follows:

\begin{itemize}
\setlength{\itemsep}{2pt}
\setlength{\parsep}{0pt}
\setlength{\parskip}{0pt}
\item We introduce the box-driven class-wise masking (BCM) model to help remove the irrelevant regions of each class. It also provides an obvious hint of the foreground region, which could directly contribute to the segmentation learning.
\item Filling rate guided adaptive loss (FR-Loss) is proposed to help select the most confident locations in the score map for back propagation and ignore the wrongly labeled pixels in the proposals.
\item Extensive experiments on PASCAL VOC 2012 benchmark demonstrate that the proposed method is effective and achieves the state-of-the-art results.

\end{itemize}

\begin{figure*}[t]
	\begin{center}
		\includegraphics[width=1\linewidth]{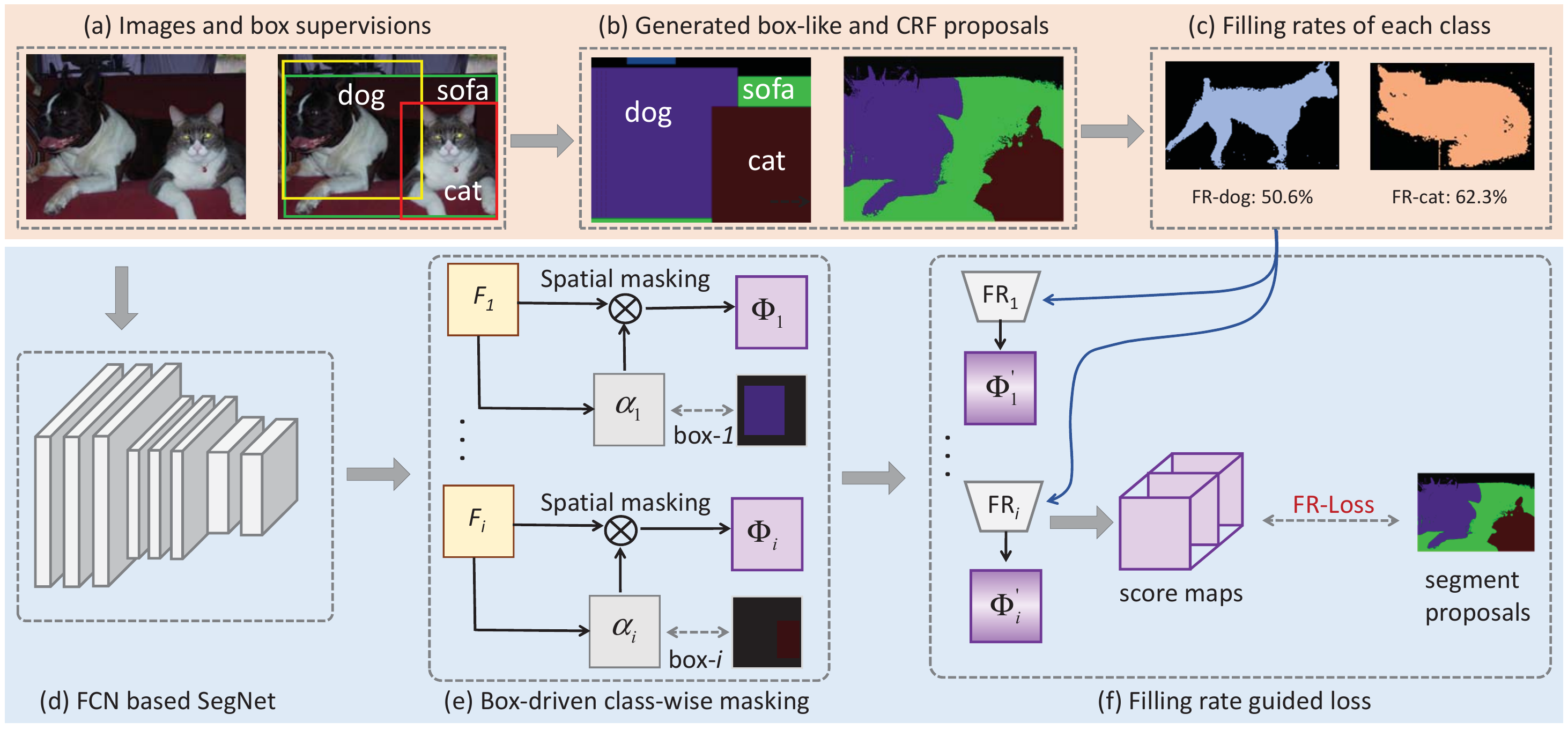}
	\end{center}
	\caption{Pipeline of the proposed method. For a given image and its corresponding bounding boxes (a), we first generate the rectangle annotations (Box-like) and apply the unsupervised CRF \cite{R9} to generate segment proposals (b). We then calculate the mean filling rates of each class (c) with the CRF proposals and their corresponding boxes. With the images and segment proposals, we train the FCN based model (d), e.g., the DeepLab-LargeFOV network \cite{R17}. We add a box-driven class-wise masking (BCM) model (e) to generate class-aware masks via segmentation learning with box-like labels. The learned masks can implement spatial masking on the features of each class, separately. For each forward step, we rank the scores of each class in the prediction layer and adopt the filling rate guided loss (FR-loss) (f) to select the most confident locations for back propagation and ignore the weak ones. FR-loss could reduce the negative effects caused by the wrongly labeled pixels in the proposals.}
	\label{fig:fig2}
%\vspace{-3.0mm}
\end{figure*}

\section{Related Work}
In this section, we briefly introduce the fully and weakly supervised semantic segmentation methods which are related to our work.
%\vspace{-3.0mm}
\subsection{Fully Supervised Semantic Segmentation}
Fully supervised semantic segmentation has achieved a series of progress \cite{R1, R3, R4, R24, R25, R26, R28, R36}, among which \cite{R1} is the first to introduce the Fully Convolutional Neural Networks (FCN) structure into segmentation field. Following this work, a large number of improvements \cite{R2, R16, R17, R21, R27, R34, R47, R48, R50, R51, R52, R53} through redesigning or adjusting the network structures have been proposed. Chen et al. \cite{R17} introduce the atrous convolution for dense prediction and enlarge the receptive field of view. Zhen et al. \cite{R17} propose to adopt the dense CRF \cite{R9} with Gaussian pairwise potentials as a Recurrent Neural Network (RNN) to refine coarse outputs from a traditional CNN. Recently, an encoder-decoder based atrous separable convolution model \cite{R34} has achieved the state-of-the-art performance for fully supervised semantic image segregation.

\subsection{Weakly Supervised Semantic Segmentation}
Recently, a large number of weakly supervised methods explore to learn semantic segmentation with supervision of image labels \cite{R39, R40, R41, R42, R54}, points \cite{R35}, scribbles \cite{R44, R45, R49}, and bounding boxes \cite{R5, R6, R7, R38}. The bounding boxes based methods are the most related works to this paper. BoxSup \cite{R5} introduces the recursive training procedure with the supervision of segment proposals. WSSL \cite{R6} proposes an expectation-maximization algorithm with segment proposals generated by the dense CRF \cite{R9}. Whereas SDI \cite{R7} tries to produce segment proposals via combining MCG \cite{R10} and GarbCut \cite{R11} methods. Li et al. \cite{R38} explore to segment the instance with both the bounding box supervision and the image tags. Different from these methods, we propose a box-driven class-wise masking (BCM) model to help remove the backgrounds before predicting the final segmentation. Unlike the global spatial attention model adopted in previous works \cite{R12, R13, R29, R31}, the proposed BCM can learn specific attention maps for each class. To our knowledge, we are the first one to introduce the mean filling rate (FR) as a stable guidance through selecting the most confident locations in the score map for back propagation. The proposed FR-loss can adaptively select the reliable pixels and ignore the wrongly labeled pixels in the pseudo proposals.

\section{Our Method}

\subsection{Overview}

We present the proposed weakly supervised semantic segmentation framework with only bounding box supervision in this section. This framework can learn semantic masks from weakly box-level annotations via the box-driven class-wise masking (BCM) model and the filling rate guided loss (FR-Loss). In the following paragraph, we first describe the general pipeline, then introduce the details of each components.

There are mainly two steps for the proposed method, as shown in Figure \ref{fig:fig2}. First, we generate the pixel-level proposals with the bounding box annotations and calculate the mean filling rates of each class. Then, we train the Fully Convolutional Network (FCN) based model with the proposed box-driven class-wise masking (BCM) model and filling rate guided loss (FR-loss).

\textbf{Proposals Generating and Filling Rates Computing.} The first step for weakly supervised semantic segmentation is to generate proper supervision labels from given bounding boxes, as shown in Figure \ref{fig:fig2} (b). The simplest yet widely used method is to convert the bounding boxes into rectangle segments directly, named as box-like proposals. Considering that the rectangle segments contain lots of wrongly labeled background regions within the bounding box, it is necessary to be further refined. There are several popular methods to generate high-quality segment proposals with bounding box labels, among which dense CRF \cite{R6}, MCG \cite{R10} and GrabCut \cite{R11} are the mostly used approaches. For fair comparison with the baseline model \cite{R6}, we choose the same unsupervised dense CRF as the default option to generate proposals. With the CRF proposals and their corresponding boxes, we can calculate the mean filling rates of each class, as shown in Figure \ref{fig:fig2} (c).

\textbf{Model Training with BCM and FR-loss.} As shown in Figure \ref{fig:fig2} (d), the backbone model in this paper is DeepLab-LargeFOV model \cite{R17}. Similar with the original FCN \cite{R1} training procedure, we also initialize this model with a VGG-16 model \cite{R15} pre-trained on ImageNet \cite{R20}. This backbone model is comparable with the ones used in the compared methods \cite{R5, R6, R7}. The FCN model takes the images as its inputs and the segment proposals as the supervision. To this end, the FCN model can be trained in an end-to-end manner. Note that in our case the quality of the supervision information in weakly supervised task is not guaranteed, so we add a box-driven class-wise masking (BCM) model to generate class-aware masks via segmentation learning with the box-like labels. The learned masks can implement spatial masking on the features of each class, separately. For each forward step, we rank the scores of each class in the prediction layer and adopt the filling rate guided loss (FR-loss) to select the most confident locations for back propagation and ignore the weak ones. FR-loss could reduce the negative effects caused by the wrongly labeled pixels in proposals. Details of them will be described in the next two sub-sections.

\subsection{Box-driven Class-wise Masking}

To remove the irrelevant regions in the feature maps, we need to learn specific masking maps for each class. Thus we design a box-driven class-wise masking (BCM) model to guide the learning of the segmentation model. We apply the masking on the FC-7 layer (note: implemented by convolution) of VGG-16 model \cite{R15} to mask the irrelevant regions. As shown in Figure \ref{fig:fig2} (e), the output features of FCN based SegNet are evenly sliced into $N$ branches, corresponding to the $N$ classes. For each branch, we add an binary attention model to produce a weights map for masking. To give a clear hint, we introduce the box-like mask to guide the attention map via adding a Mean Squared Error (MSE) loss on pixels of the attention map $\alpha_c$ and its corresponding mask $M_c$ of class-$c$
\begin{equation}
L_{bcm(c)}  = \sum\limits_{h = 1}^H {\sum\limits_{w = 1}^W {\left\| {M_{c(h,w)}  - \alpha _{c(h,w)} } \right\|_2^2 } }
\end{equation}
where $\alpha_c$ has a size of $(H,W)$. In the similar way, the $N$ binary segmentation models can be trained separately. Then the $N$ attention maps could carry out spatial-wise masking across their corresponding feature branches. We denote $\alpha _c$ and $F_c$ as the learnt attention map and feature branch of class-$c$, respectively. Therefore, the weighted feature of class-$c$ can be denoted as
\begin{equation}
\Phi _c  = F_c  \otimes \alpha _c
\end{equation}
where $\otimes$ means the spatial-wise masking operation. Then we combine the output features of $N$ branches to produce the score map for final segmentation.

Unlike the global spatial attention model adopted in previous works \cite{R12, R13}, the proposed class-wise masking model can learn specific attention maps for each class. It contributes to the segmentation models in three respects: 1) It can remove the irrelevant regions in the feature maps, such as the backgrounds. 2) It can learn $N$ specific masking maps to fit each class, which may differ greatly from each other in shapes and sizes. 3) As the mask is learnt under the supervision of bounding box, thus it could provide a clear object hint for the segmentation learning.

\subsection{Filling Rate Guided Adaptive Loss}

Above box-driven class-wise masking model can guide the FCN to learn foreground features softly, we further explore to improve the segmentation learning in this subsection. Note that the wrongly labeled regions of the pixel-level proposals have negative effects on model training, recognizing the negative regions will be helpful. A possible solution is to ignore the pixels with small confident values in the score map, which may be the wrongly labeled pixels. In the weakly supervised mode, there are no guaranteed pixel-level annotations like the fully supervised mode, thus it is hard to determine how much percentage of pixels to be ignored. To address this problem, we introduce the filling rate guided adaptive loss (FR-loss). We intuitively find that the percentage of foreground pixels within the bounding box should be similar for the same class. Whereas the pixel filling ratios of two classes are usually different. Therefore, we first calculate the mean pixel filling rates of each class with pixel-level proposals and their corresponding boxes. For a given class-$c$, we denote the number of foreground pixels in the $i$-$th$ proposal and box as $P_{proposal} (i)$ and $P_{box} (i)$, respectively. Then the mean filling rate of class-$c$ can be defined as

\begin{equation}
FR_c  = \frac{1}{{N_c }}\sum\nolimits_{i = 1}^{N_c } {\frac{{P_{proposal} (i)}}{{P_{box} (i)}}}
\end{equation}
where $N_c$ means the number of bounding boxes in class-$c$. Therefore, the mean filling rate of each class can be used to determine how much percent of the most confident pixels can be left for training or being ignored. In this way, the segmentation loss could be adjusted by the filling rates of each class. The FR-loss for one sample can be denoted as

\begin{equation}
L_{fr}  = \sum\nolimits_{{\rm{c}} = 1}^N {\sum\nolimits_{i = 1}^{top (FR_c)} {L_c (i)} }
\end{equation}
where $L_c (i)$ means the loss of the $i$-$th$ pixel with class-$c$, and the super-parameter $top$ is determined by the mean filling rate of each class. This loss guides the score map to learn the most confident regions adaptively.

\begin{figure}[t]
	\begin{center}
		\includegraphics[width=0.9\linewidth]{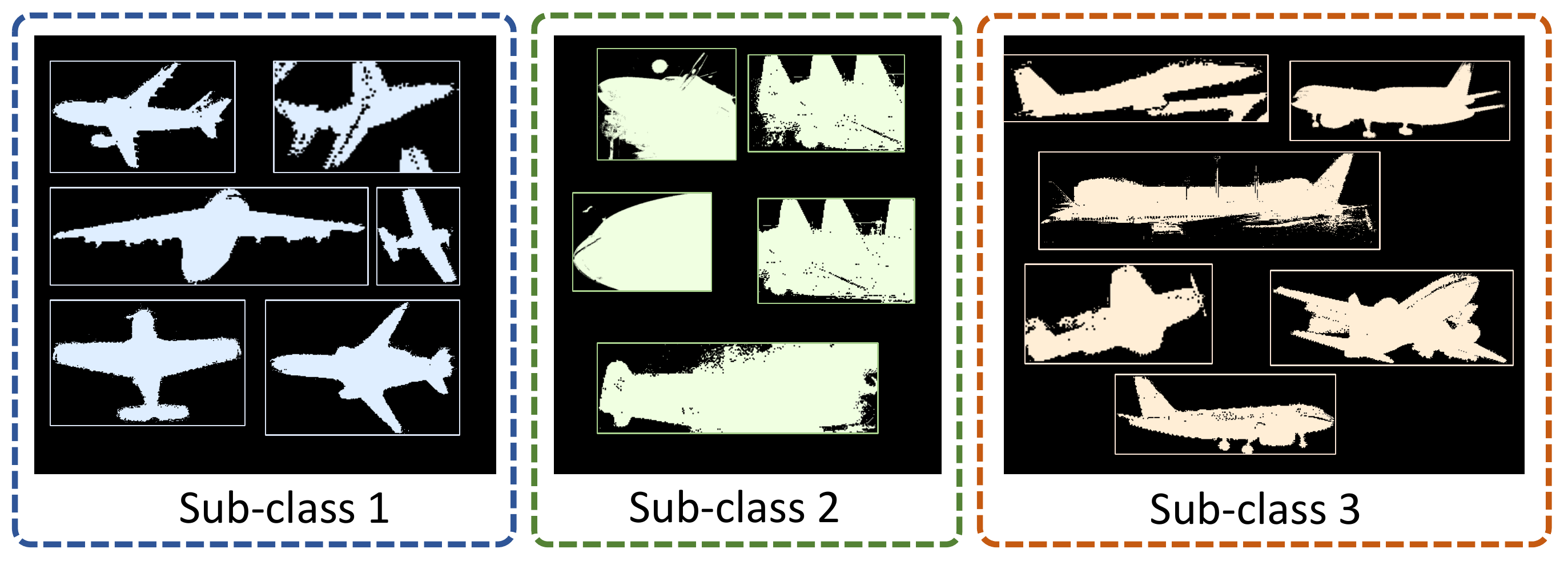}
	\end{center}
	\caption{Examples of three sub-classes from airplane class. It is obvious that the middle sub-class has a larger filling rate than the right and left sub-classes. The mean filling rates of each sub-class could better represent different kinds of sample in one class.}
	\label{fig:fig3}
\vspace{-3.0mm}
\end{figure}

\textbf{Refine the Filling Rates with Sub-class Clustering.} Considering the situation that two objects from the same class may have different filling rates due to the shape and pose varieties, we try to refine the filling rates via k-meas clustering method \cite{R46} to classify each class into several sub-classes. As shown in Figure \ref{fig:fig3}, we show the examples of three clustered sub-classes of airplane. Visually, three sub-classes are reasonable which can better represent three groups of boxes. Thus we take the mean filling rates of each sub-class to refine the FR-loss. In this situation, the FR-loss for one sample can be denoted as

\begin{equation}
L_{fr}  = \sum\nolimits_{c = 1}^N {\sum\nolimits_{sc}^3 {\sum\nolimits_{i = 1}^{top(FR_{(c,sc)} )} L _{(c,sc)} } } (i)
\end{equation}
where $L_{(c,sc)} (i)$ means the loss of the $i$-$th$ pixel with class-$c$ and sub-class-$sc$. Note that $L_{(c,sc)}(i)$ is 0 when this pixel does not belong to this sub-class.

In retrospect, the class-wise masking model introduced in last subsection and the FR-loss can work together to guide the segmentation learning in a `soft' manner, achieving comparable performance with the full-supervised model. The overall loss for one sample can be denoted as

\begin{equation}
L_{all}  = L_{fr}  + \lambda  \cdot \sum\nolimits_{c = 1}^N {L_{bcm(c)} }
\end{equation}
where $\lambda$  is the hypermeter which is set to 0.01 in our experiments, $N$ is the number of classes. We will evaluate the proposed methods in experiments.

\section{Experiments}

In the experiments, we first evaluate the effectiveness of our method on the Pascal VOC 2012 semantic segmentation dataset \cite{R14}, then compare the proposed method with three state-of-the-art methods under weakly supervision and semi-supervision conditions separately.

\begin{table}[t]
\tabcolsep=15.5pt
\footnotesize
\centering
\begin{tabular}{|l|l|c|}\hline
\textbf{Methods}       &\textbf{Units} &\textbf{mIoU}  \\
\hline\hline
Baseline \cite{R6}  &-  &60.6 \\
\hline\hline

\multirow{7}{*}{Ours}
&CM          &63.4\\
&BGM          &64.9\\
&\textbf{BCM}          &65.6\\\cline{2-3}

&Global-loss          &64.1\\
&FR-loss       &65.8\\
&\textbf{FR-loss(Refine)}        &66.3\\\cline{2-3}
&\textbf{BCM + FR-loss(Refine)}       &\textbf{66.8} \\
\hline
\end{tabular}
\caption{Evaluate the effectivenesses of BCM and FR-loss on VOC2012 validation set. All models are based on the same Deeplab VGG16-LargeFOV backbones. The performance is evaluated in terms of mean IoU ($\%$). CM: class-wise masking without box supervision, BGM: box-driven global masking, Global-loss: all boxes adopt the same global filling rate of 0.6.}
\label{tab:tab1}
\vspace{-2.0mm}
\end{table}

\begin{figure}[t]
	\begin{center}
		\includegraphics[width=0.95\linewidth]{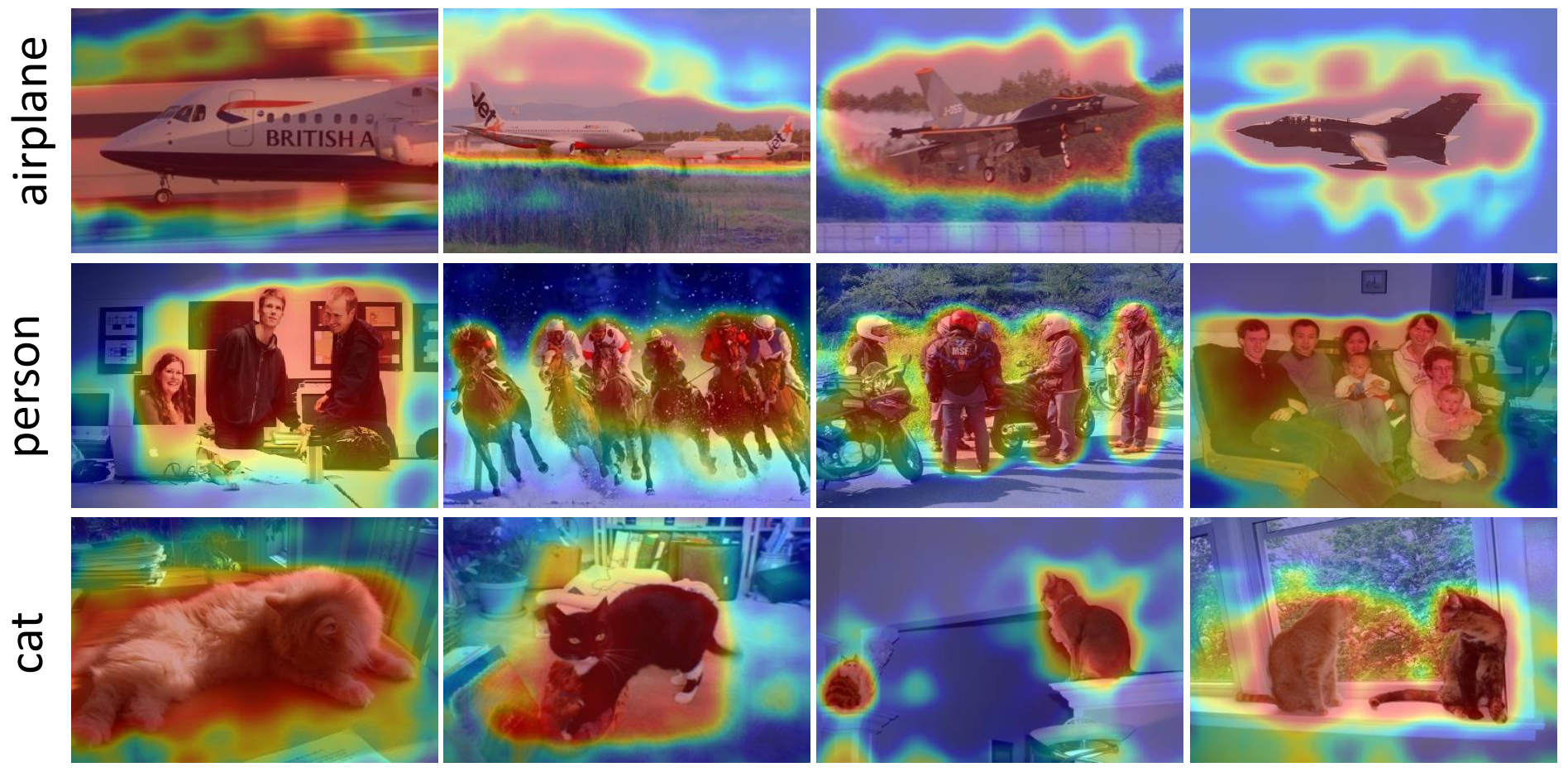}
	\end{center}
\vspace{-3.0mm}
	\caption{Visualization of the BCM learnt masking maps. It shows that most of the backgrounds are removed.}
	\label{fig:fig4}
%\vspace{-3.0mm}
\end{figure}

\subsection{Experimental Setup}

\textbf{Dataset.} We evaluate the proposed framework on the widely used Pascal VOC2012 segmentation benchmark \cite{R14}. It contains 21 classes with pixel-level annotations. There are 1,464 images in the training set and 1,449 in the validation set, and the left 1,456 images are for testing. Following the same setting in the compared methods \cite{R5, R6, R7}, we augment the training set with the data from SBD \cite{R19}. Consequently, there are 10,582 images in the training set and 1,449 images in the validation set. We train our model with the augmented training set and test it on the validation set to compare with other methods.

\textbf{Implementation Details.} We adopt the publicly released and widely used DeepLab-LargeFOV \cite{R17} model as the backbone network. It is based on a VGG-16 \cite{R15} network which has been pre-trained on ImageNet \cite{R20}. We train the proposed model under several different supervision settings. We first train the Deeplab-largeFOV model with the rectangle-box supervision. Further, we change the segments supervision into CRF-Box segments for finetuning, and regard it as the baseline model. Based on above model, we train the models with the proposed Box-driven Class-wise Masking (BCM) model and the Filling Rate guided Loss (FR-loss). We train the baseline model with roughly 20k iterations, and further fine-tune them with/without the BCM and  FR-loss for 5k more iterations. In addition, we also evaluate the performance in the semi-supervised condition through adding 1,449 samples with ground-truth labels. The initial learning rates of the above models are 0.001 and decreased by 10 times after every 3k iterations, with a mini-batch size of 16/20 for the model with/without BCM. We take SGD as the default optimizer. For all the training phases, only flipping and cropping are adopted for data augmentation. With the well-trained FCN models, we can predict the semantic masks for the given images. Note the forward-passes of masking layers in BCM are parallel, the forward-passing time are very close to the baseline model, i.e., 42.7ms vs. 39.3ms per image. We also implement the dense-CRF \cite{R9} for post-processing on the masks. We adopt the same parameters of the dense-CRF with the compared work \cite{R6}. All experiments are implemented on a Nvidia TitanX GPU platform with the Caffe \cite{R32} framework.

\textbf{Evaluation Metrics and Compared Methods.} We adopt the ``comp6'' protocol to evaluate the performance. The accuracy is reported in terms of mean pixel Intersection-over-Union (mean IoU). We compare with three start-of-the-art methods (i.e., BoxSup \cite{R5}, WSSL \cite{R6} and SDI \cite{R7}) on VOC 2012 dataset under both weakly supervised and semi-supervised conditions with bounding box annotations.

\begin{table}[t]
\tabcolsep=7.5pt
\footnotesize
\centering
\begin{tabular}{|c|c|c|l|c|}\hline

\textbf{Modes}           &$\#$ GT       &$\#$ Box           &\textbf{Methods}              &\textbf{mIoU}\\
\hline\hline
\multirow{8}*{Weak}   &\multirow{8}*{-}     &\multirow{8}*{10,582}                &BoxSup$_{Box}$ \cite{R5}     & 52.3\\\cline{4-5}
                      &                     &                                     &WSSL$_{Box}$ \cite{R6}       & 52.5\\\cline{4-5}
                      &                     &                                     &SDI$_{Box}$ \cite{R7}        & 61.2\\\cline{4-5}
                      &                     &                                    &\textbf{Ours$_{Box}$}     & 54.9\\\cline{4-5}
                      &                     &                                   &BoxSup$_{MCG}$ \cite{R5}     & 62.0\\\cline{4-5}
                      &                     &                                   &WSSL$_{CRF}$ \cite{R6}       & 60.6\\\cline{4-5}
                      &                     &                                    &SDI$_{M+G}$ \cite{R7}    & 65.7\\\cline{4-5}
                      &                     &                                  &\textbf{Ours$_{CRF}$}     & \textbf{66.8}\\
\hline\hline

\multirow{5}*{Semi}   &\multirow{5}*{1,464}     &\multirow{5}*{9,118}                &WSSL$_{Box}$ \cite{R6}     & 62.1\\\cline{4-5}
                      &                     &                                    &BoxSup$_{MCG}$ \cite{R5}      & 63.5\\\cline{4-5}
                      &                     &                                  &WSSL$_{CRF}$        & 65.1\\\cline{4-5}
                      &                     &                                &SDI$_{M+G}$ \cite{R7}     & 65.8\\\cline{4-5}
                      &                     &                                  &\textbf{Ours$_{CRF}$}     & \textbf{67.5}\\
\hline\hline
Full   &10,582  &-                &DeepLab-LargeFOV \cite{R17} &69.8 \\
\hline
\end{tabular}
\caption{Weakly and Semi-supervised results on VOC2012 validation set. With only 1/10 labeled segments, our method can achieve comparable performance with the fully supervised model. Box: directly using rectangle proposals, M+G: using the combined labels with both MCG and GrabCut.}
%\vspace{-2.0mm}
\label{tab:tab2}
\end{table}

\begin{table}[t]
\tabcolsep=7pt
\footnotesize
\centering
\begin{tabular}{|c|c|c|l|c|}\hline
\textbf{Modes}           &$\#$ GT       &$\#$ Box           &\textbf{Methods}              &\textbf{mIoU}\\
\hline\hline
\multirow{2}*{Weak}   &\multirow{2}*{-}     &\multirow{2}*{10,582}                &SDI \cite{R7}        & 69.4\\\cline{4-5}
                       &                     &                                   &\textbf{Ours}     & \textbf{70.2}\\
\hline\hline

Semi   &1,464     &9,118               &\textbf{Ours}     & \textbf{71.6}\\

\hline\hline
Full   &10,582  &-                 &DeepLab-ResNet-101 \cite{R17}  &74.5\\
\hline
\end{tabular}
\caption{Results of ResNet-101 backbone on VOC2012 validation set. Our method outperforms the compared SDI \cite{R7} method, achieving comparable performance with the fully supervised one.}
\label{tab:tab3}
%\vspace{-2.0mm}
\end{table}

\subsection{Effectiveness of BCM and FR-Loss}
We first evaluate the proposed framework with BCM and FR-loss, the results are shown in Table \ref{tab:tab1}. Based on the Deeplab-LargeFOV model and CRF-box proposals, finetuning with the BCM model or the FR-loss can achieve 65.6$\%$ or 65.8$\%$ mean IoU accuracy, respectively. Both of them outperform the baseline model with obvious margins. When the BCM and FR-loss work together, we achieve 66.8$\%$ accuracy. The results show that the proposed BCM and FR-loss are effective and jointly combing the two modules can further enhance the performance. We also evaluate several variants of the proposed BCM and FR-loss. Experimental results show that the box-driven class-wise masking model performs better than the global one (BGM). We show the BCM learnt masks in Figure \ref{fig:fig4}. Without the influence of the cluttered backgrounds, the segmentation learning could be more stable. It demonstrates that the class-wise attention model can guide the FCN model to learn more effective features, and the filling rate guided adaptive loss can help reduce the negative effects from the wrongly labeled proposals.

\begin{figure*}[t]
	\begin{center}
		\includegraphics[width=0.95\linewidth]{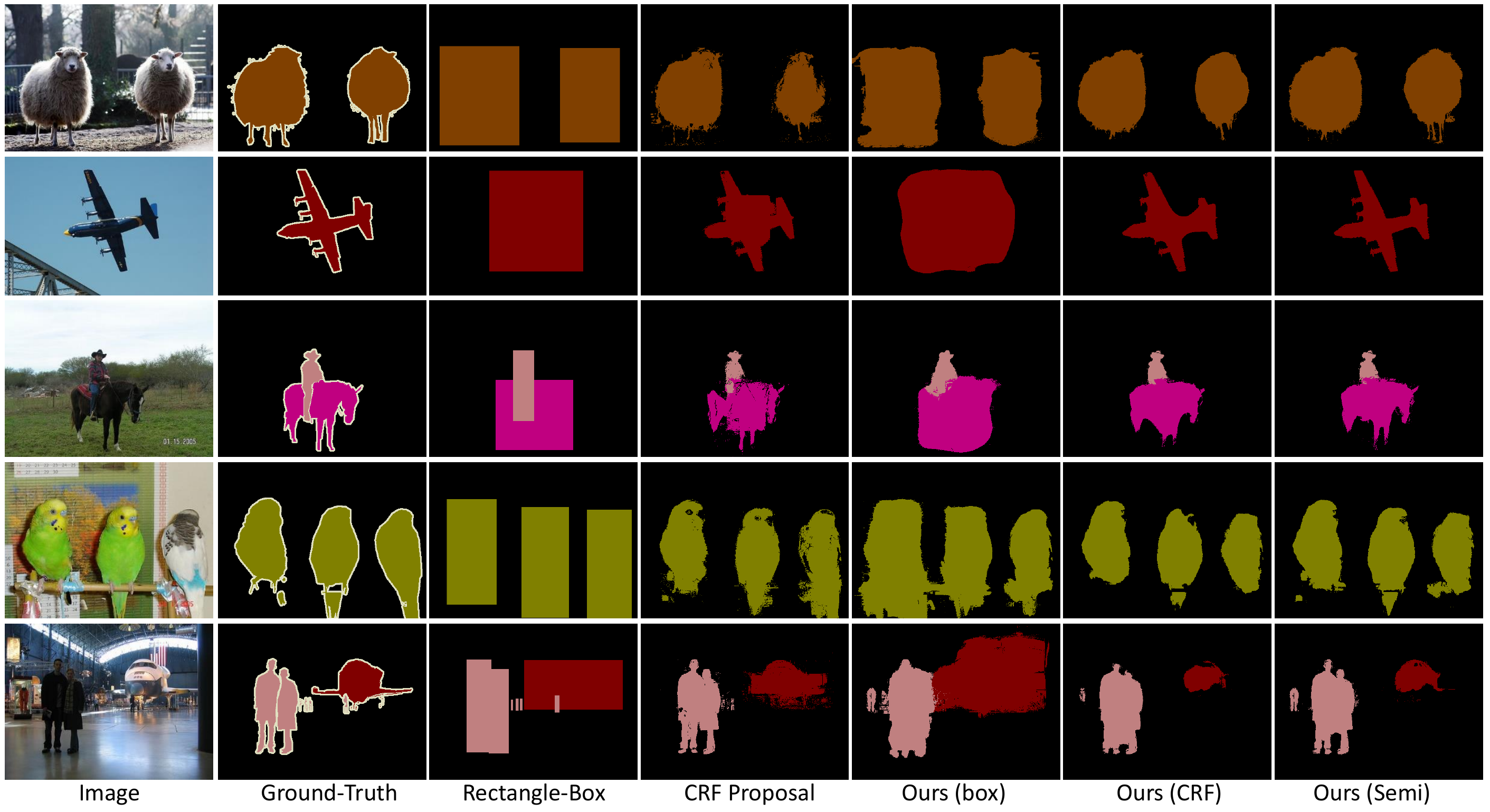}
	\end{center}
\vspace{-3.0mm}
	\caption{Examples of the segmentation results with proposed method. Original images are in the first column. The second column is the ground-truth segmentations. The 3-rd and 4-th columns are rectangle-box and CRF proposals. The following two columns show the results trained with rectangle-box and CRF proposals, respectively. The last column shows the results of semi-supervised model.}
	\label{fig:fig5}
%\vspace{-3.0mm}
\end{figure*}

\begin{table*}[t]
\tabcolsep=2.2pt
\footnotesize
\centering
\begin{tabular}%
{|c|c|c|c|c|c|c|c|c|c|c|c|c|c|c|c|c|c|c|c|c|c|c|}\hline

\textbf{Methods}       &bkg  &aero &bike &bird &boat &bottle  &bus  &car  &cat &chair &cow &table &dog &horse &moto &person &plant &sheep &sofa &train &tv  &\textbf{mean}\\
\hline\hline
Weak(box)    &78.3 &37.4 &20.6 &46.6 &44.9 &64.5 &80.7 &68.1 &59.8 &32.5 &65.7 &58.4 &61.6 &51.2 &53.2 &60.5   &47.5  &60.0  &49.3 &64.2  &49.4 &\textbf{54.9}\\
\hline
Weak(CRF)     &89.8 &68.3 &27.1 &73.7 &56.4 &72.6 &84.2 &75.6 &79.9 &35.2 &78.3 &53.2 &77.6 &66.4 &68.1 &73.1 &56.8 &80.1 &45.1 & 74.7 &54.6 &\textbf{66.8}\\
\hline
Semi          &90.4 &72.3 &27.5 &76.1 &57.8 &72.4 &85.6 &76.6 &81.3 &35.9 &80.2 &53.0 &78.4 &68.2 &69.7 &73.9 &58.1 &82.1 &45.3 &76.5 &57.0  &\textbf{67.5}\\
\hline
\end{tabular}
\caption{Per class results of our method on VOC2012 validation set. The performance is evaluated in terms of mean IoU ($\%$).}
\label{tab:tab4}
\vspace{-2.0mm}
\end{table*}

\begin{figure}[t]
	\begin{center}
		\includegraphics[width=0.8\linewidth]{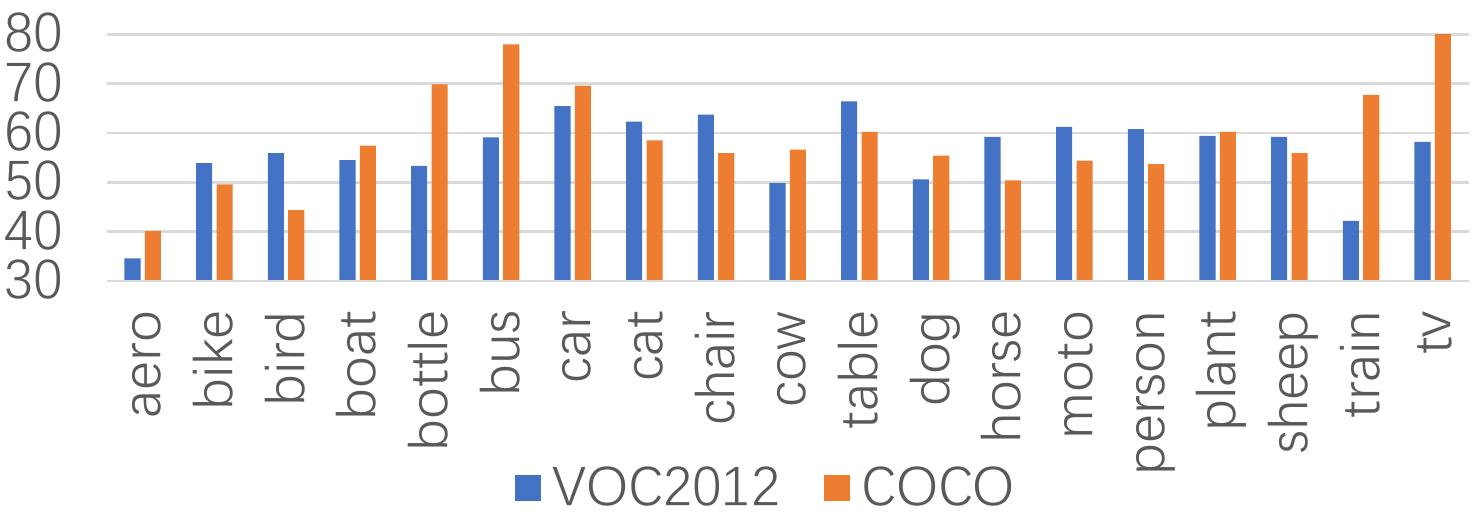}
	\end{center}
\vspace{-1.0mm}
	\caption{Filling rates of each class on VOC2012 and COCO. The filling rates are calculated with the generated pixel-level proposals. It is obvious that the filling rate can serve as an important cue for adjusting the pseudo labels. }
	\label{fig:fig_fr}
\vspace{-3.0mm}
\end{figure}

\subsection{Comparison with the State-of-the-art Methods}
We compare with three state-of-the-art methods, i.e., BoxSup \cite{R5}, WSSL \cite{R6} and SDI \cite{R7}.

\textbf{Results of Weakly-supervised Conditions.} We first compare the results under the weakly supervised condition, as shown in Table \ref{tab:tab2}. In this case, the only supervision label is the bounding box. We compare with BoxSup \cite{R5}, WSSL \cite{R6} and SDI \cite{R7} from two aspects of view. Firstly, we compare the models trained with raw rectangle-box segments. The proposed method outperforms the BoxSup and WSSL, whereas SDI performs better which adopts an iterative training to update the segments from time to time. Secondly, we compare the models trained with the pre-processed segments. Our method outperforms all compared results and achieves an amazing performance with 66.8$\%$ mean IoU accuracy, which is very close to the full-supervised model. Note that our method adopts the same CRF-Box segments and the same base model with WSSL \cite{R6}, whereas the performance of our method exceeds WSSL by roughly 6$\%$. In addition, we compare the models trained with ResNet-101 backbone, as shown in Table \ref{tab:tab3}. We achieve 70.2$\%$ mean IoU accuracy. The results demonstrate that the proposed method is effective for learning robust and accurate representations from bounding box annotations.

\textbf{Results of Semi-supervised Conditions.} We further compare with other methods in the semi-supervised task. In this task, 1,464 ground-truth labels are added for training. Although the amount of labeled samples is small which is only 1/10 of the training sets, they help improve the performance greatly. As shown in Table \ref{tab:tab2}, the proposed method achieves 67.5$\%$ mean IoU accuracy, outperforming all the compared methods. With the extra 1/10 labeled segments, our model gets 0.7$\%$ improvement than its weakly version. The results prove that our semi-supervised model can achieve comparable performance with the fully supervised model, showing the proposed BCM and FR-loss are still effective in semi-supervised mode.

\begin{figure}[t]
	\begin{center}
		\includegraphics[width=1\linewidth]{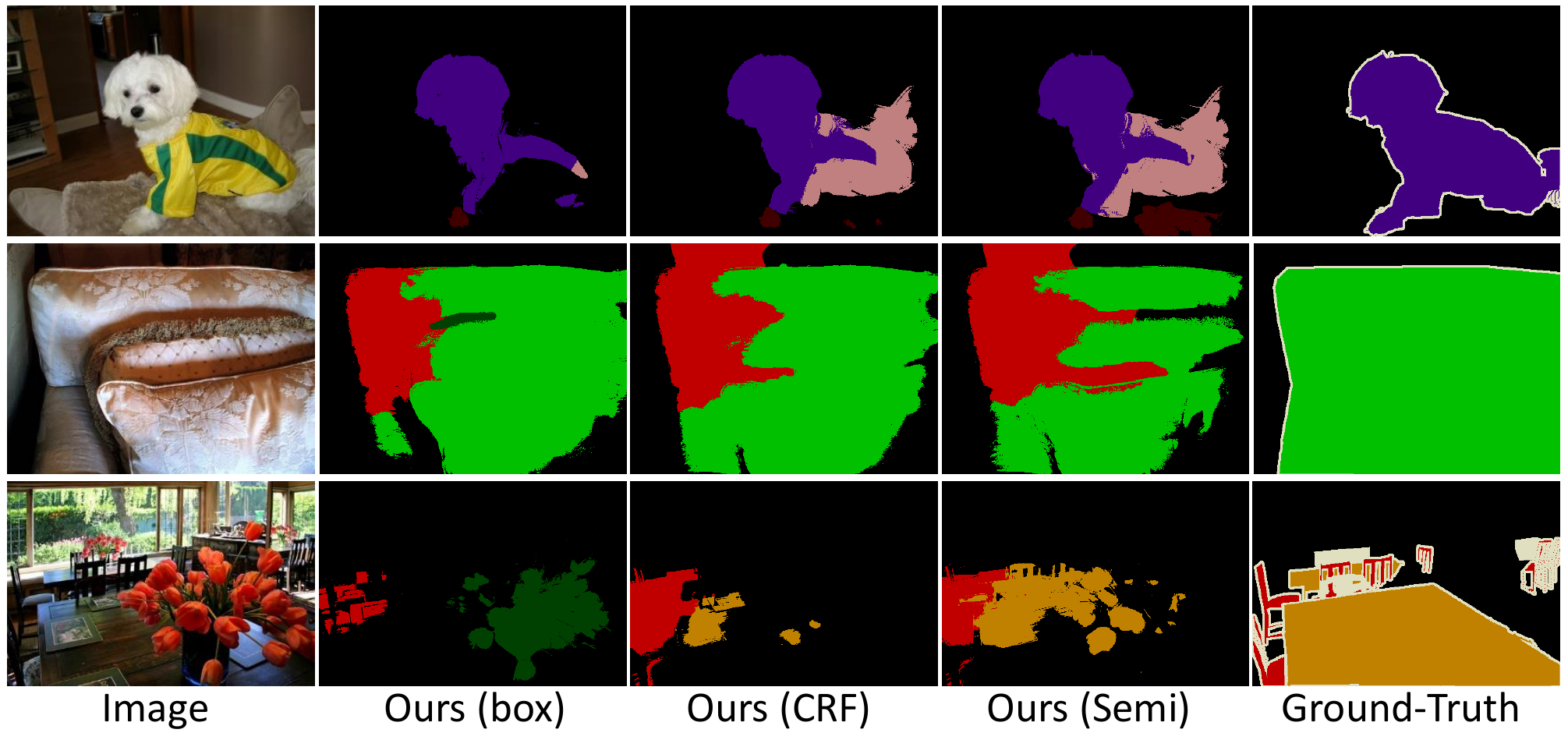}
	\end{center}
\vspace{-1.0mm}
	\caption{Failure examples of proposed method. Though our model achieves satisfying performances under both weakly and semi-supervised conditions, there are some frustrated results. For example, a dog wearing a cloth in the first image makes the model confused.}
	\label{fig:fig6}
\vspace{-3.0mm}
\end{figure}

\subsection{Discussions}
Above results have shown that the proposed method can learn better segmentations than the compared methods. To provide a comprehensive analysis, we report the per class results of proposed models, as shown in Table \ref{tab:tab4}. We also calculate the per class FR of VOC2012 and COCO \cite{R33}, as shown in Figure \ref{fig:fig_fr}. It shows that the filling rates of VOC2012 and COCO are basically consistent, besides several classes, e.g., \emph{train} and \emph{tv}. It is obvious that among 21 classes, \emph{airplane} and \emph{sheep} are easy to segment, whereas \emph{person} and \emph{chair} are difficult. This result is consistent with the qualitative results shown in Figure \ref{fig:fig5} and \ref{fig:fig6}. The generated CRF proposals can help the model learn pixel-level representations, achieving satisfied results. In addition, with the help of proposed BCM and FR-Loss, the model can reach a comparable performance with the fully supervised model. There are also some hard examples which bring great challenges to weakly supervised methods. As shown in Figure \ref{fig:fig6}, it is very difficult for the model trained with limited and weakly labeled data to distinguish the classes in chaotic and complex scenes. This problem is worth to be deeply studied in the future work. Here, we will discuss the proposed methods separately.

\textbf{Box-driven Class-wise Masking.} With the class-level supervision, soft attention model based methods are widely adopted to guide the CNN model to learn better representations. Generally, the learned attention map usually contains object shape information. However, the global attention map can not learn multiple accurate shape templates for each class at the same time. In our method, the class-wise masking model can solve this problem. As shown in Figure \ref{fig:fig4}, the learnt masks can remove the irrelevant regions and cluttered backgrounds to effectively contribute to the segmentation learning. In brief, BCM is helpful for box-driven weakly supervised segmentation through effective masking.

\textbf{Filling Rate Guided Adaptive Loss.} The FR-loss can guide the segmentation model to learn object masks in a soft manner, reducing the negative impacts from the wrongly labeled proposals. In this paper, we first directly set the mean filling rates of each class as the default value for determining the most confident locations. FR can be regarded as a kind of prior knowledge which could supervise the weakly learning procedure. Note that the filling rate of a class is independent of the others. The FR-loss is still effective when several classes have similar FR values and will not affect the performance. Considering that some samples may be greatly different from other samples though with the same class, the strategy of choosing top scores could be further improved. Thus we refine the filling rate via clustering each class into several sub-classes. It will be interesting to explore a better way to classify the sub-classes. We leave this problem as our future work.

\section{Conclusion}
In this paper, we have introduced a Box-driven Class-wise Masking (BCM) model to learn attention maps of each class. It can produce class-aware attentive maps for segmentation task learning, and provide an obvious hint whether this box or region contains a specific class. Moreover, based on the region-level segment proposals generated from the bounding boxes, we have proposed a Filling Rate guided adaptive loss (FR-loss) to help the model ignore the wrongly labeled pixels in proposals. FR-loss can adjust the model learning with global statistical information. The proposed BCM and FR-loss can work together to help reduce the negative impacts from wrongly labeled proposals. We evaluate the proposed method on the challenging PASCAL VOC 2012 benchmark and compare with other methods. Extensive experimental results show that the proposed method is effective and achieves the state-of-the-art results. In future, we will explore the jointly learning of the object detection and segmentation tasks to find more positive interactions between them.

\section*{Acknowledgement}
This work is jointly supported by National Key Research and Development Program of China (2016YFB1001000), National Natural Science Foundation of China (61525306, 61633021, 61721004, 61420106015, 61806194), Capital Science and Technology Leading Talent Training Project (Z181100006318030), and Beijing Science and Technology Project (Z181100008918010). This work is also supported by grants from NVIDIA and the NVIDIA DGX-1 AI Supercomputer.

{\small
\bibliographystyle{ieee_fullname}
\bibliography{mybibfile}

\begin{thebibliography}{10}\itemsep=-1pt

\bibitem{R42}
Jiwoon Ahn and Suha Kwak.
\newblock Learning pixel-level semantic affinity with image-level supervision
  for weakly supervised semantic segmentation.
\newblock In {\em CVPR}, 2018.

\bibitem{R35}
Amy Bearman, Olga Russakovsky, Vittorio Ferrari, and Li Fei-Fei.
\newblock What's the point: Semantic segmentation with point supervision.
\newblock In {\em ECCV}, 2016.

\bibitem{R51}
Kai Chen, Jiangmiao Pang, Jiaqi Wang, Yu Xiong, Xiaoxiao Li, Shuyang Sun,
  Wansen Feng, Ziwei Liu, Jianping Shi, Wanli Ouyang, Chen~Change Loy, and
  Dahua Lin.
\newblock Hybrid task cascade for instance segmentation.
\newblock {\em arXiv preprint arXiv:1901.07518}, 2019.

\bibitem{R12}
Long Chen, Hanwang Zhang, Jun Xiao, Liqiang Nie, Jian Shao, and Tat-Seng Chua.
\newblock Sca-cnn: Spatial and channel-wise attention in convolutional networks
  for image captioning.
\newblock {\em arXiv preprint arXiv:1611.05594}, 2016.

\bibitem{R17}
Liang-Chieh Chen, George Papandreou, Iasonas Kokkinos, Kevin Murphy, and Alan~L
  Yuille.
\newblock Semantic image segmentation with deep convolutional nets and fully
  connected crfs.
\newblock In {\em ICLR}, 2015.

\bibitem{R13}
Liang-Chieh Chen, Yi Yang, Jiang Wang, Wei Xu, and Alan~L Yuille.
\newblock Attention to scale: Scale-aware semantic image segmentation.
\newblock In {\em CVPR}, 2016.

\bibitem{R34}
Liang-Chieh Chen, Yukun Zhu, George Papandreou, Florian Schroff, and Hartwig
  Adam.
\newblock Encoder-decoder with atrous separable convolution for semantic image
  segmentation.
\newblock {\em arXiv preprint arXiv:1802.02611}, 2018.

\bibitem{R24}
Jifeng Dai, Kaiming He, Yi Li, Shaoqing Ren, and Jian Sun.
\newblock Instance-sensitive fully convolutional networks.
\newblock In {\em ECCV}, 2016.

\bibitem{R5}
Jifeng Dai, Kaiming He, and Jian Sun.
\newblock Boxsup: Exploiting bounding boxes to supervise convolutional networks
  for semantic segmentation.
\newblock In {\em ICCV}, 2015.

\bibitem{R52}
Xu Dan, Wanli Ouyang, Xiaogang Wang, and Nicu Sebe.
\newblock Pad-net: Multi-tasks guided prediction-and-distillation network for
  simultaneous depth estimation and scene parsing.
\newblock In {\em CVPR}, 2018.

\bibitem{R48}
Henghui Ding, Xudong Jiang, Bing Shuai, Ai Qun~Liu, and Gang Wang.
\newblock Context contrasted feature and gated multi-scale aggregation for
  scene segmentation.
\newblock In {\em CVPR}, 2018.

\bibitem{R14}
Mark Everingham, SM~Ali Eslami, Luc Van~Gool, Christopher~KI Williams, John
  Winn, and Andrew Zisserman.
\newblock The pascal visual object classes challenge: A retrospective.
\newblock {\em IJCV}, 2015.

\bibitem{R54}
Junsong Fan, Zhaoxiang Zhang, and Tieniu Tan.
\newblock Cian: Cross-image affinity net for weakly supervised semantic
  segmentation.
\newblock {\em arXiv preprint arXiv:1811.10842}, 2018.

\bibitem{R47}
Zhang Hang, Kristin Dana, Jianping Shi, Zhongyue Zhang, Xiaogang Wang, Ambrish
  Tyagi, and Amit Agrawal.
\newblock Context encoding for semantic segmentation.
\newblock In {\em CVPR}, 2018.

\bibitem{R31}
Kota Hara, Ming-Yu Liu, Oncel Tuzel, and Amir-massoud Farahmand.
\newblock Attentional network for visual object detection.
\newblock {\em arXiv preprint arXiv:1702.01478}, 2017.

\bibitem{R19}
Bharath Hariharan, Pablo Arbel{\'a}ez, Lubomir Bourdev, Subhransu Maji, and
  Jitendra Malik.
\newblock Semantic contours from inverse detectors.
\newblock In {\em ICCV}, 2011.

\bibitem{R36}
Kaiming He, Georgia Gkioxari, Piotr Doll{\'a}r, and Ross Girshick.
\newblock Mask r-cnn.
\newblock In {\em ICCV}, 2017.

\bibitem{R16}
Kaiming He, Xiangyu Zhang, Shaoqing Ren, and Jian Sun.
\newblock Deep residual learning for image recognition.
\newblock In {\em CVPR}, 2016.

\bibitem{R40}
Zilong Huang, Xinggang Wang, Jiasi Wang, Wenyu Liu, and Jingdong Wang.
\newblock Weakly-supervised semantic segmentation network with deep seeded
  region growing.
\newblock In {\em CVPR}, 2018.

\bibitem{R32}
Yangqing Jia, Evan Shelhamer, Jeff Donahue, Sergey Karayev, Jonathan Long, Ross
  Girshick, Sergio Guadarrama, and Trevor Darrell.
\newblock Caffe: Convolutional architecture for fast feature embedding.
\newblock In {\em ACM ICM}, 2014.

\bibitem{R7}
Anna Khoreva, Rodrigo Benenson, Jan Hosang, Matthias Hein, and Bernt Schiele.
\newblock Simple does it: Weakly supervised instance and semantic segmentation.
\newblock In {\em CVPR}, 2017.

\bibitem{R9}
Philipp Kr{\"a}henb{\"u}hl and Vladlen Koltun.
\newblock Efficient inference in fully connected crfs with gaussian edge
  potentials.
\newblock In {\em NeurIPS}, 2011.

\bibitem{R38}
Qizhu Li, Anurag Arnab, and Philip~HS Torr.
\newblock Weakly-and semi-supervised panoptic segmentation.
\newblock In {\em ECCV}, 2018.

\bibitem{R45}
Di Lin, Jifeng Dai, Jiaya Jia, Kaiming He, and Jian Sun.
\newblock Scribblesup: Scribble-supervised convolutional networks for semantic
  segmentation.
\newblock In {\em CVPR}, 2016.

\bibitem{R25}
Guosheng Lin, Chunhua Shen, Anton van~den Hengel, and Ian Reid.
\newblock Efficient piecewise training of deep structured models for semantic
  segmentation.
\newblock In {\em CVPR}, 2016.

\bibitem{R33}
Tsung-Yi Lin, Michael Maire, Serge Belongie, James Hays, Pietro Perona, Deva
  Ramanan, Piotr Doll{\'a}r, and C~Lawrence Zitnick.
\newblock Microsoft coco: Common objects in context.
\newblock In {\em ECCV}, 2014.

\bibitem{R1}
Jonathan Long, Evan Shelhamer, and Trevor Darrell.
\newblock Fully convolutional networks for semantic segmentation.
\newblock In {\em CVPR}, 2015.

\bibitem{R49}
Tang Meng, Federico Perazzi, Abdelaziz Djelouah, Ismail~Ben Ayed, Christopher
  Schroers, and Yuri Boykov.
\newblock On regularized losses for weakly-supervised cnn segmentation.
\newblock In {\em ECCV}, 2018.

\bibitem{R6}
George Papandreou, Liang-Chieh Chen, Kevin Murphy, and Alan~L Yuille.
\newblock Weakly-and semi-supervised learning of a dcnn for semantic image
  segmentation.
\newblock In {\em ICCV}, 2015.

\bibitem{R26}
Pedro Pinheiro and Ronan Collobert.
\newblock Recurrent convolutional neural networks for scene labeling.
\newblock In {\em ICML}, 2014.

\bibitem{R21}
Pedro~O Pinheiro, Ronan Collobert, and Piotr Dollar.
\newblock Learning to segment object candidates.
\newblock In {\em NeurIPS}, 2015.

\bibitem{R10}
Jordi Pont-Tuset, Pablo Arbelaez, Jonathan~T Barron, Ferran Marques, and
  Jitendra Malik.
\newblock Multiscale combinatorial grouping for image segmentation and object
  proposal generation.
\newblock {\em IEEE TPAMI}, 2017.

\bibitem{R37}
Carolina Redondo-Cabrera and Roberto~J L{\'o}pez-Sastre.
\newblock Learning to exploit the prior network knowledge for weakly-supervised
  semantic segmentation.
\newblock {\em arXiv preprint arXiv:1804.04882}, 2018.

\bibitem{R11}
Carsten Rother, Vladimir Kolmogorov, and Andrew Blake.
\newblock Grabcut: Interactive foreground extraction using iterated graph cuts.
\newblock {\em ACM Transactions on Graphics}, 2004.

\bibitem{R20}
Olga Russakovsky, Jia Deng, Hao Su, Jonathan Krause, Sanjeev Satheesh, Sean Ma,
  Zhiheng Huang, Andrej Karpathy, Aditya Khosla, Michael Bernstein, et~al.
\newblock Imagenet large scale visual recognition challenge.
\newblock {\em IJCV}, 2015.

\bibitem{R15}
Karen Simonyan and Andrew Zisserman.
\newblock Very deep convolutional networks for large-scale image recognition.
\newblock {\em arXiv preprint arXiv:1409.1556}, 2014.

\bibitem{R50}
Chunfeng Song, Yan Huang, Wanli Ouyang, and Liang Wang.
\newblock Mask-guided contrastive attention model for person re-identification.
\newblock In {\em CVPR}, 2018.

\bibitem{R3}
Chunfeng Song, Yongzhen Huang, Zhenyu Wang, and Liang Wang.
\newblock 1000fps human segmentation with deep convolutional neural networks.
\newblock In {\em ACPR}, 2015.

\bibitem{R44}
Paul Vernaza and Manmohan Chandraker.
\newblock Learning random-walk label propagation for weakly-supervised semantic
  segmentation.
\newblock In {\em CVPR}, 2017.

\bibitem{R46}
Kiri Wagstaff, Claire Cardie, Seth Rogers, and Stefan Schr{\"o}dl.
\newblock Constrained k-means clustering with background knowledge.
\newblock In {\em ICML}, 2001.

\bibitem{R39}
Xiang Wang, Shaodi You, Xi Li, and Huimin Ma.
\newblock Weakly-supervised semantic segmentation by iteratively mining common
  object features.
\newblock In {\em CVPR}, 2018.

\bibitem{R41}
Yunchao Wei, Huaxin Xiao, Honghui Shi, Zequn Jie, Jiashi Feng, and Thomas~S
  Huang.
\newblock Revisiting dilated convolution: A simple approach for weakly-and
  semi-supervised semantic segmentation.
\newblock In {\em CVPR}, 2018.

\bibitem{R4}
Zifeng Wu, Yongzhen Huang, Yinan Yu, Liang Wang, and Tieniu Tan.
\newblock Early hierarchical contexts learned by convolutional networks for
  image segmentation.
\newblock In {\em ICPR}, 2014.

\bibitem{R2}
Zifeng Wu, Chunhua Shen, and Anton van~den Hengel.
\newblock Wider or deeper: Revisiting the resnet model for visual recognition.
\newblock {\em arXiv preprint arXiv:1611.10080}, 2016.

\bibitem{R53}
Dan Xu, Wanli Ouyang, Xavier Alamedapineda, Elisa Ricci, Xiaogang Wang, and
  Nicu Sebe.
\newblock Learning deep structured multi-scale features using attention-gated
  crfs for contour prediction.
\newblock In {\em NeurIPS}, 2017.

\bibitem{R29}
Kelvin Xu, Jimmy Ba, Ryan Kiros, Kyunghyun Cho, Aaron Courville, Ruslan
  Salakhudinov, Rich Zemel, and Yoshua Bengio.
\newblock Show, attend and tell: Neural image caption generation with visual
  attention.
\newblock In {\em ICML}, 2015.

\bibitem{R27}
Fisher Yu and Vladlen Koltun.
\newblock Multi-scale context aggregation by dilated convolutions.
\newblock {\em arXiv preprint arXiv:1511.07122}, 2015.

\bibitem{R28}
Shuai Zheng, Sadeep Jayasumana, Bernardino Romera-Paredes, Vibhav Vineet,
  Zhizhong Su, Dalong Du, Chang Huang, and Philip~HS Torr.
\newblock Conditional random fields as recurrent neural networks.
\newblock In {\em ICCV}, 2015.

\end{thebibliography}
}

\end{document}